\documentclass{article}

\usepackage{microtype}
\usepackage{graphicx}
\usepackage{subfigure}
\usepackage{booktabs} % for professional tables
\usepackage{makecell}
\usepackage{multirow}
\usepackage{comment}
\usepackage{relsize}
\usepackage{hyperref}

\usepackage[accepted]{icml2024}

\usepackage{amsmath}
\usepackage{amssymb}
\usepackage{mathtools}
\usepackage{amsthm}

\usepackage[capitalize,noabbrev]{cleveref}

\theoremstyle{plain}

\theoremstyle{definition}

\theoremstyle{remark}

\usepackage[textsize=tiny]{todonotes}

\icmltitlerunning{{TimeHF: Billion-Scale Time Series Models Guided by Human Feedback}}

\begin{document}

\twocolumn[
\icmltitle{{TimeHF: Billion-Scale \underline{Time} Series Models Guided by \underline{H}uman \underline{F}eedback}}




\icmlsetsymbol{equal}{*}
\begin{icmlauthorlist}
\icmlauthor{Yongzhi Qi}{jd}
\icmlauthor{Hao Hu}{jd}
\icmlauthor{Dazhou Lei}{bjtu}
\icmlauthor{Jianshen Zhang}{jd}
\icmlauthor{Zhengxin Shi}{jd}
\icmlauthor{Yulin Huang}{jd}
\icmlauthor{Zhengyu Chen}{jd}
\icmlauthor{Xiaoming Lin}{jd}
\icmlauthor{Zuo-Jun Max Shen}{berkeley,hku}
\end{icmlauthorlist}

\icmlaffiliation{jd}{JD.com, Beijing, China}
\icmlaffiliation{bjtu}{Beijing Jiaotong University, Beijing, China}
\icmlaffiliation{berkeley}{College of Engineering, UC Berkeley, Berkeley, CA 94720, USA}
\icmlaffiliation{hku}{Faculty of Engineering, Faculty of Business and Economics, University of Hong Kong, Hong Kong, China}

\icmlcorrespondingauthor{Dazhou Lei}{dzlei@bjtu.edu.cn}

\icmlcorrespondingauthor{Zuo-Jun Max Shen}{maxshen@berkeley.edu}
\icmlkeywords{Machine Learning, Artificial Intelligence}

\vskip 0.3in
]



\printAffiliationsAndNotice{}

\begin{abstract}

Time series neural networks perform exceptionally well in real-world applications but encounter challenges such as limited scalability, poor generalization, and suboptimal zero-shot performance. Inspired by large language models, there's interest in developing large time series models (LTM) to address these issues. However, current methods struggle with training complexity, adapting human feedback, and achieving high predictive accuracy. We introduce TimeHF, a novel pipeline for creating LTMs with 6 billion parameters, incorporating human feedback. We use patch convolutional embedding to capture long time series information and design a human feedback mechanism called time-series policy optimization. Deployed in JD.com’s supply chain, TimeHF handles automated replenishment for over 20,000 products, improving prediction accuracy by 33.21\% over existing methods. This work advances LTM technology and shows significant industrial benefits.
\end{abstract}

\section{Introduction}
\label{intro}

Benefiting from the success of large models in fields such as language processing \citep{dubey2024llama} and computer vision \citep{kirillov2023segment}, large-scale time series models (LTM) have rapidly developed in recent years. 

\begin{figure}[ht]
\begin{center}
\centerline{\includegraphics[width=\columnwidth]{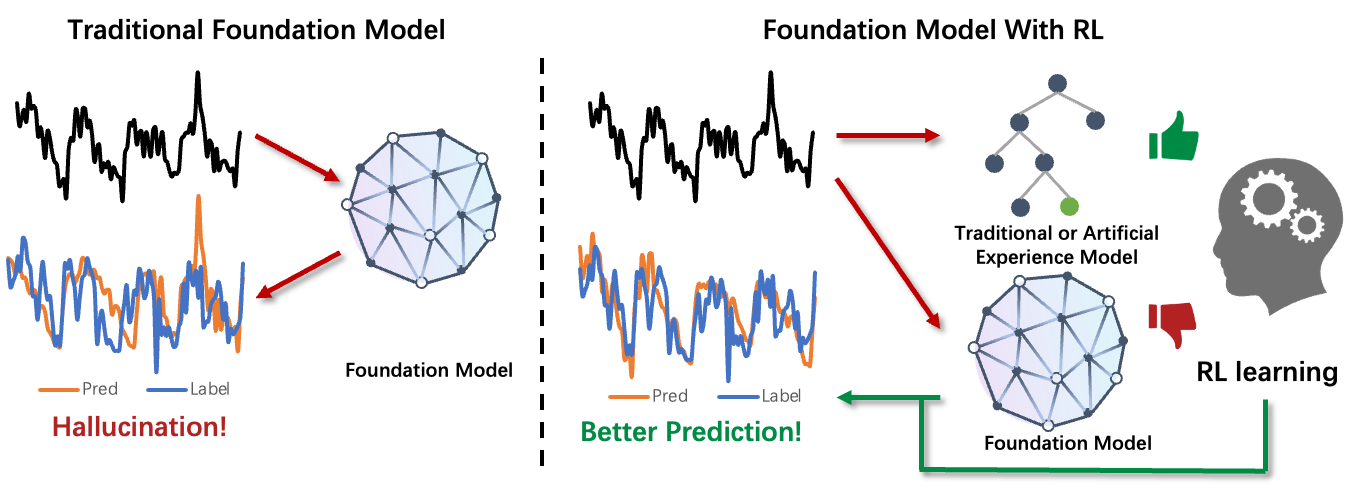}}
\caption{\textbf{Left:} Traditional time series foundation models, while performing well in common scenarios, may be overly sensitive to noises in training datasets, resulting in hallucinations in complex scenarios. \textbf{Right:} With RLHF, feedback contrast pairs—constructed by interpretable small models created by human experts—guide the model to gradually shift toward more accurate predictions.}
\label{fig_overview}
\end{center}
\vskip -0.2in
\end{figure}

Roughly, two typical strategies can be adopted to construct LTM: one involves designing appropriate prompts and tokenizers to discretize and align the time-series modality with the textual modality, followed by leveraging pre-trained large language models (LLMs) for inference \citep{zhou2023one}; the other is based on the transformer architecture, formulating a pure LTM from scratch \citep{woo2024unified}. 

Scaling laws state that increasing the size of models and training datasets typically leads to performance improvements, which have been widely validated in language and vision domains \citep{alabdulmohsin2022revisiting}. However, existing publicly available time series datasets generally suffer from limited data volume and strong regularity, small models can already effectively capture these regular patterns, making it difficult for scaling laws to take effect. Currently, the largest pure time series model contains only 710 million parameters \citep{ansari2024chronos}, significantly smaller than the large language models which reach up to 405 billion parameters \citep{dubey2024llama}. 

Although LTM may perform well in some common scenarios, time series can exhibit various patterns or distributions across different fields. In a specific domain such as forecasting sales of a long-tail product, prediction models must be adjusted or tuned to mitigate hallucinations \citep{xu2024hallucination}, preventing inaccurate predictions or the emergence of outliers. Popular tuning methods in the LTM domain include Supervised Fine-Tuning (SFT) \citep{liutimer} and Retrieval-Augmented Generation (RAG) \citep{yang2024timerag}. To align with human preferences, Reinforcement Learning from Human Feedback (RLHF) \citep{knox2011augmenting} directly optimizes model behavior using human feedback. RLHF has demonstrated superior performance in complex tasks and dynamic environments and has been widely applied in fields such as text generation \citep{ouyang2022training}. 

In this paper, we propose a novel framework for building LTMs to address the aforementioned challenges (see Figure \ref{fig_overview} for an overview). Leveraging JD.com's extensive sales data, we construct a high-quality time series dataset that spans a variety of scenarios, including standard products, new products, seasonal items, bestsellers, long-tail items, and intermittent items. This diverse dataset significantly enhances the richness and complexity of training data, thereby enabling the development of the LTMs of greater scale. We train PCLTMs with 300M, 1.6B, and 6B parameters, zero-shot experiments demonstrate their superior performance. 

To further effectively incorporate human expertise into LTMs, we propose the first RLHF approach for time series prediction models, termed Timeseries policy optimization (TPO). It utilizes specialized predictive models (small models) built by human experts as proxies for expert knowledge and cognition to generate prediction pairs reflecting diverse human preferences in complex time series scenarios. Compared to state-of-the-art models (GPT4TS), our framework achieves an average reduction in Mean Squared Error (MSE) of 7.09\%. Our contributions are summarized as follows:

\begin{itemize}
    \item We propose a standard for constructing large-scale, high-quality time series datasets, including data augmentation such as high-dimensional aggregated and interpretable component prediction data, data balancing, diversity ranking, etc.
    Based on this standard, we construct a large-scale time series dataset of 210B data points with intricate and diverse temporal patterns.
    
    \item  To address the intricate cross-dependencies in complex, long time series, we introduce a novel patch-based convolutional large time series model (TimeHF).
    Based on this approach, We propose the first billion-scale pure time series model and successfully achieve industrial deployment in JD.com.
    
    \item We propose Timeseries policy optimization (TPO), the first RLHF framework applicable to time series forecasting, which enables LTMs to learn the tacit knowledge of experienced time-series experts. Numerical studies show that our framework can improve the zero-shot capabilities of LTMs significantly.
\end{itemize}

\begin{table}[h]
\label{tab_model_comp}
\caption{Comparison of Our Method with Existing LTMs}
\begin{center}
\setlength{\tabcolsep}{1pt}
\begin{scriptsize} 
\begin{sc}
\begin{tabular}{@{}llcccccc@{}}
\toprule
Category      & Method    & \makecell{Training \\ Token} & \makecell{Trainable \\ Parameter} & \makecell{Zero-Shot \\ Forecast} & SFT       & RLHF      \\ 
\midrule
LLM-based  & TimeLLM   & -              & 6M                  & $\surd$         & $\surd$ &           \\
             & GPT4TS    & -              & 4M                  & $\surd$         & $\surd$ &           \\ 
\midrule
Pure LTM   & Chronos   & 84B            & 710M                & $\surd$         & $\surd$ &           \\
             & Moirai    & 28B            & 311M                & $\surd$         & -          &           \\
             & Timer     & 28B            & 67M                 & $\surd$         & $\surd$ &           \\
             & MOMENT    & 1B             & 385M                & $\surd$         & $\surd$ &           \\
             & TimesFM   & 100B           & 200M                & $\surd$         & $\surd$ &           \\
             & \textbf{TimeHF}      & \textbf{210B}           & \textbf{6B}                  & $\surd$         & $\surd$ & $\surd$ \\
\bottomrule
\end{tabular}
\end{sc}
\end{scriptsize}
\end{center}
\vskip -0.1in
\end{table}

\section{Related Work}

\subsection{Pretraining Dataset in LTMs}

Table \ref{tab_model_comp} provides a summary of the key differences between recent LTMs, they can be broadly categorized into two types. The first type relies on foundational models from the language or image domains, assuming inherent similarities in time series data \citep{gruver2024large, zhou2023one, jin2023time, rasul2023lag, ansari2024chronos, chen2024visionts}, requiring additional data transformation modules. The second type, including TimesFM \citep{das2023decoder}, MoiRAI \citep{woo2024unified}, and Timer \citep{liutimer}. These models can extract deeper insights from time series data, but they all treat patches as independent segments, limiting the model's ability to capture inter-patch information. 

Pre-training data are crucial for LTMs, but currently available public time series datasets are insufficient in richness and scale, and existing construction methods are simplistic, lacking emphasis on data quality. These issues may lead to insufficient diversity and balance in the data. 

Common datasets for LTMs include TSLib \citep{wu2022timesnet}, the Monash Database \citep{godahewa2021monash}, the GluonTs library \citep{alexandrov2020gluonts}, and the LOTSA dataset \citep{woo2024unified}. However, these publicly available datasets are far smaller than those used for LLMs. For example, LOTSA integrates various accessible datasets but contains only 27B observations, while TimesFM uses a mixed dataset of 100B observations. While TimesFM’s synthetic data reflects common patterns, it sacrifices some authenticity. Our pretraining dataset combines public datasets with JD.com sales data, leveraging aggregation, predictive synthesis, and advanced data balancing and processing techniques.

\subsection{Reinforcement Learning from Human Feedback}
SFT and RLHF are two popular strategies to enhance LLMs accuracy and adaptability for specific tasks. Compared to SFT, RLHF achieves better results with less data \citep{christiano2017deep, stiennon2020learning, ouyang2022training}. Many studies enhance reinforcement learning strategies in RLHF for LLMs, such as Proximal Policy Optimization (PPO) and direct Preference Optimization (DPO) \citep{schulman2017proximal, rafailov2024direct, ahmadian2024back}. Recognizing the critical role of RLHF in optimizing LLMs, we propose a specialized RLHF framework TPO, tailored for pure LTMs. This framework significantly improves both effectiveness and efficiency compared to traditional reinforcement learning methods. Table \ref{tab_rl_models} compares TPO with other reinforcement learning approaches.

\begin{table*}[h!]
\caption{Comparison of TPO \& Other Reinforcement Learning Models}
\label{tab_rl_models}
\begin{center}
\begin{small}
\begin{sc}
\begin{tabular}{ccccccc}
\toprule
Model & Actor Model & Critic Model & Reward Model  & Online Learning & Efficiency \\
\midrule
PPO          & Required        & Required         & Required         & $\surd$                              & Low            \\
RLOO         & Required        & Not Required     & Required         & $\surd$                              & Medium         \\
DPO          & Required        & Not Required     & Not Required     & $\times$                              & High           \\
\textbf{TPO}          & Required    & Not Required & Not Required & $\surd$                    & High       \\
\bottomrule
\end{tabular}
\end{sc}
\end{small}
\end{center}
\vskip -0.1in
\end{table*}

\section{Method}
\begin{figure*}[ht]
\begin{center}
\centerline{\includegraphics[width=2\columnwidth]{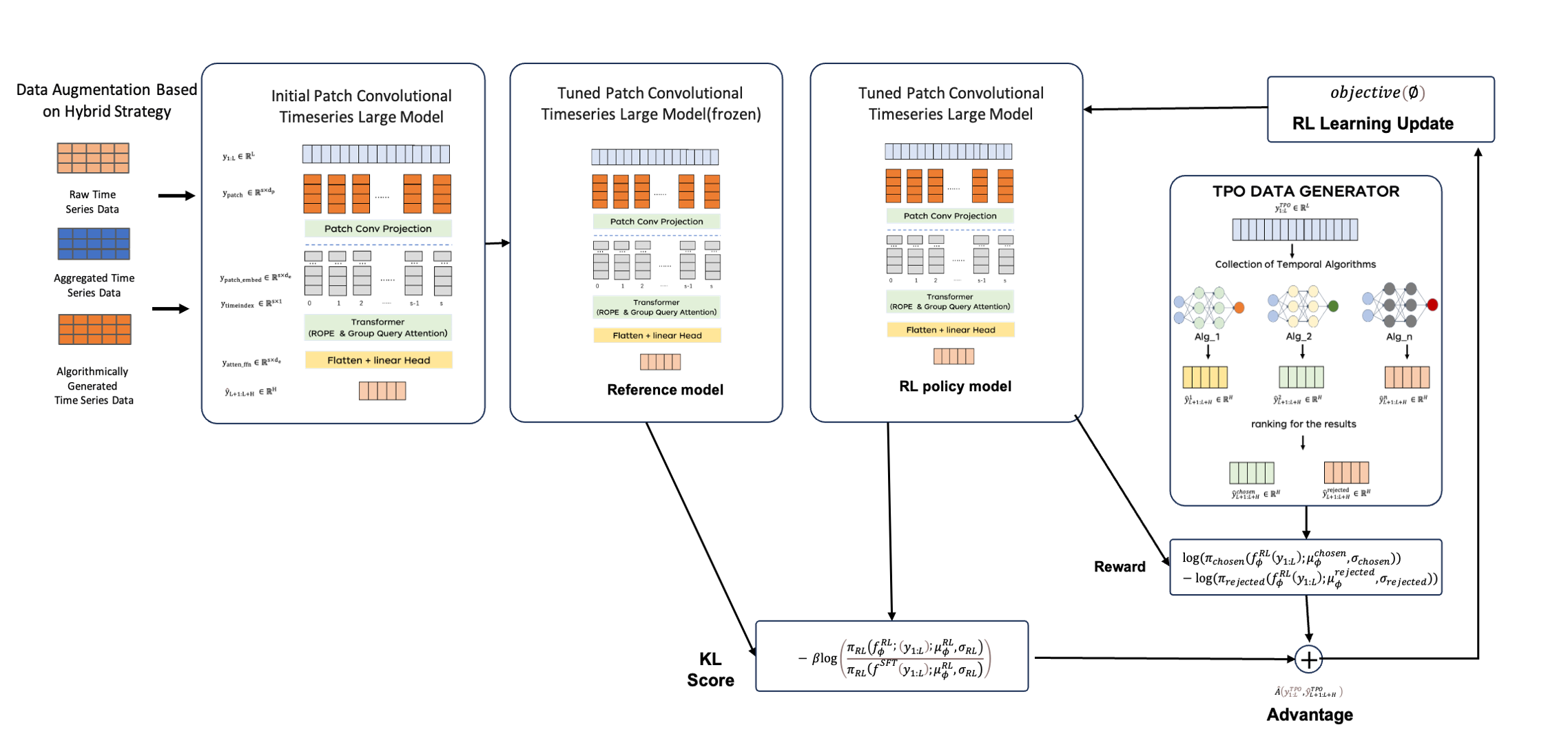}}
\caption{Model architecture of our method.}
\label{fig_method}
\end{center}
\end{figure*}

Inspired by the widely adopted training procedures of LLMs, we design a three-step framework for LTMs: (1) base model training, (2) supervised fine-tuning, (3) reinforcement learning from human feedback. Figure \ref{fig_method} shows the whole framework of our method.

\subsection{Base Model: Patch Convolutional Large Timeseries  Model}

The proposed Patch Convolutional Large Timeseries  Model (PCLTM) employs a patch-based approach (potentially overlapping patches) and utilizes a masked encoder architecture for time series modeling. The core Transformer module follows an encoder-only architecture, incorporating various advancements from the state-of-the-art LLM architectures. We divide the input into patches and project them into vector representations. To capture intricate inter-patch information across different channels, we design a network based on convolutional layers. Further, we employ Rotary Position Embedding (ROPE) for temporal position encoding, and Grouped Query Attention (GQA), a group attention mechanism with temporal position encoding to facilitate better embedding. During training, we focus on prediction for each time point. Typical point prediction loss such as Mean Squared Error (MSE) can be utilized to update the gradients. 

\subsubsection{Cross-Patch Projection Layers}
Conventional patch encoding methods typically use linear mapping where each embedding vector only represents the information within the patch itself. Time series data, however, often has long-term dependencies, requiring stronger handling of cross-patch information. Therefore, we aim to incorporate information from a broader temporal window—beyond just the data within a single patch—right at the embedding stage. We design a convolution-based network, called patch conv module to achieve this end.

Suppose the patch length is $d_p$ and input is divided into $s$ patches. An input sequence $y_{1:L} \in \mathbb{R}^L$ after patching can be represented as:
\begin{equation}
    y_{patch} \in \mathbb{R}^{s \times d_p}.
\end{equation}

\begin{figure}[h]
\begin{center}
\centerline{\includegraphics[width=0.78\columnwidth]
{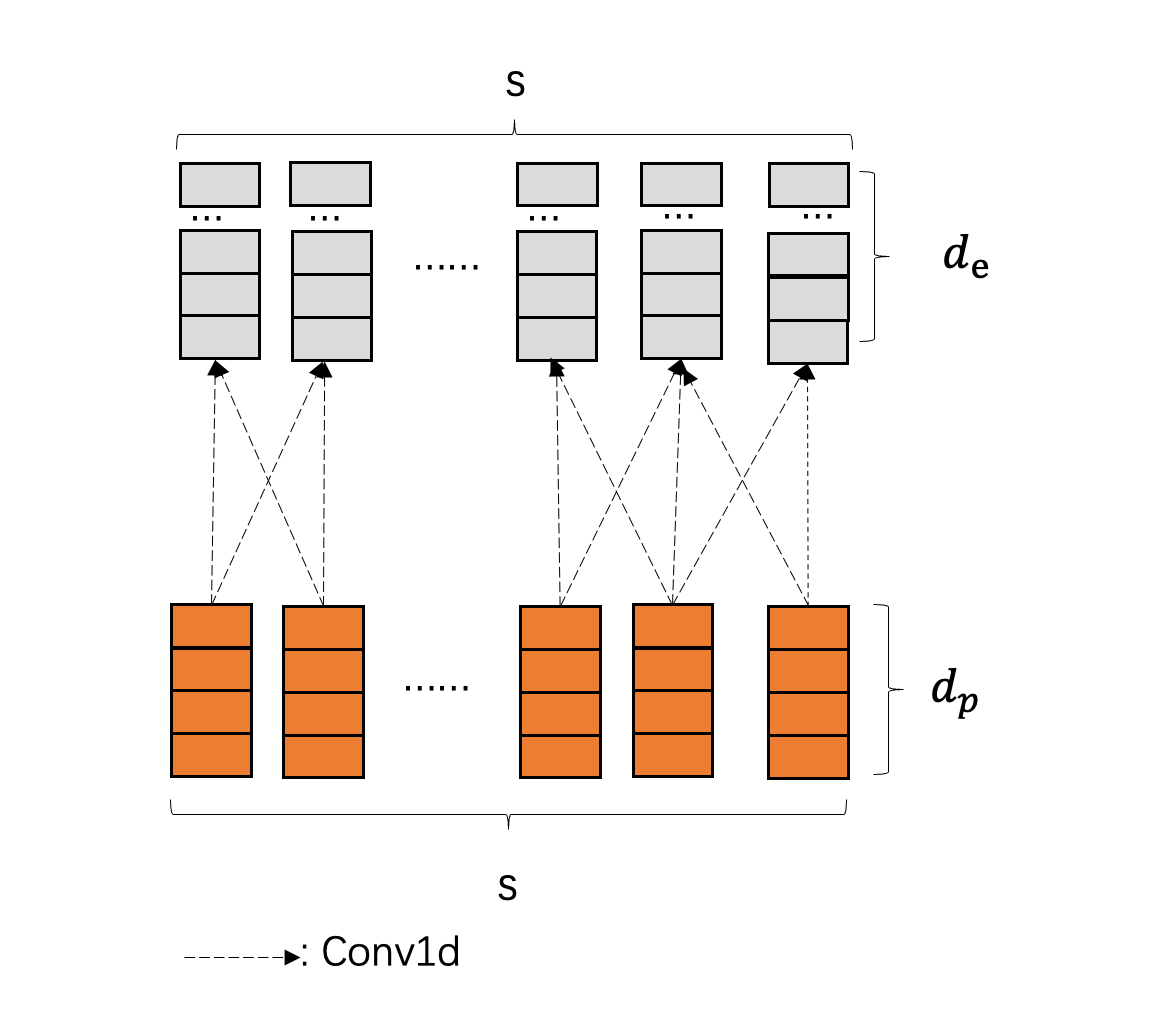}}
\caption{Cross-patch convolutional projection layers.}
\label{fig_patchconv}
\end{center}
\end{figure}

Then we encode all patches by convolutional layers with cross-patch channels. This requires transforming from $\mathbb{R}^{d_p}$ to $\mathbb{R}^{d_e}$, implying each patch is transformed to a larger vector size incorporating temporal information of a longer period. See Figure \ref{fig_patchconv} for an illustration of the cross-patch projection.
\begin{equation}
    y_{patch\_embed} = \text{patchConv}(y_{patch}) \in \mathbb{R}^{s \times d_e}
\end{equation}

\subsubsection{Group Attention Mechanism with Temporal Position Encoding}
We employ GQA to reduce the burdensome computation cost of vast amounts of attention parameters. Let \( y_{time\_index} \in \mathbb{R}^{s \times 1} \) denote the positions of all patches. Then the query, key, and value are represented by:
\begin{equation}
\begin{aligned}
q_{i} = W_{q} y_{\text{patch}\_{\text{embed}},i} \in \mathbb{R}^{d_{e}} \\
k_{i} = W_{k} y_{\text{patch}\_{\text{embed}},i}  \in \mathbb{R}^{d_{e}} \\
v_{i} = W_{v} y_{\text{patch}\_{\text{embed}},i}  \in \mathbb{R}^{d_{e}}
\end{aligned}
\end{equation}
where \( W_{q} \), \( W_{k} \), and \( W_{v} \) are the parameters of the linear transformation layers for \( q_i \), \( k_i \), and \( v_i \) ($i=1,...,s$), respectively. 

To incorporate sequential information, we further apply ROPE to both \( q_{i} \) and \( k_{i} \):
\begin{equation}
\text{Atten}_{i,j} = \text{softmax}({q_{i}}^{T} R_{i-j} k_{j}) \in \mathbb{R}^{d_{e}},
\end{equation}
\begin{equation}
R_{i-j} = \text{ROPE}(y_{\text{time\_index}})_{i-j} \in \mathbb{R}^{d_{e} \times d_{e}}.
\end{equation}
where \( R_{i-j} \) is the rotary encoding matrix. After a Feed-Forward Network (FFN) which is a common module in transformers, the output of the transformer is:
\begin{equation}
y_{\text{atten\_ffn}} = \text{FFN}(\text{Atten} * v) \in \mathbb{R}^{s \times d_{e}}.
\end{equation}

\subsubsection{Output Layers}
For the final output, an additional mapping module is needed for proper output dimensions, i.e., \(\mathbb{R}^{s \times d_e} \rightarrow \mathbb{R}^H\). We adopt a flattening operation and then use an MLP for a prediction of length $H$:
\begin{equation}
\hat{y}_{L+1:L+H} = \text{MLP}\left(flatten\left(y_{atten\_ffn}\right)\right) \in \mathbb{R}^H.
\end{equation}

\subsection{Supervised Fine-Tuning}
SFT is one of the most common methods for enhancing the performance of large models. Due to the diversity of downstream tasks, pre-training datasets typically contain a large amount of time series data in various domains. For a specific scenario, collecting domain-specific datasets and applying SFT techniques can significantly improve prediction accuracy. We also construct a bunch of fine-tuning datasets to conduct SFT. Our SFT experiments on both proprietary and public datasets demonstrate that SFT has significant advantages in improving time-series forecasting performance. Let \(y_{1:L}^{SFT}\) be the input data for SFT, then the SFT task is given by:
\[
\hat{y}_{L+1:L+H}^{SFT} = f_\theta^{SFT}(y_{1:L}^{SFT}).
\]

\subsection{Timeseries Policy Optimization}
Reinforcement learning has been widely used for LLM fine-tuning, enabling the models to learn from human preferences and enhance their performance. The RL objective \citep{ouyang2022training}for LLMs is:
\begin{equation}
\begin{aligned}
& \text{objective}(\phi) = \\
& \mathbb{E}_{(x, y) \sim D_{\pi_{\phi}^{RL}}} \left[ r_{\theta}(x, y) - \beta \log \left( \frac{\pi_{\phi}^{RL}(y \mid x)}{\pi^{SFT}(y \mid x)} \right) \right] \\
& + \gamma \mathbb{E}_{x \sim D_{\text{pretrain}}} \left[ \log \left( \pi_{\phi}^{RL}(x) \right) \right]
\end{aligned}
\label{eq_rl}
\end{equation}
where \(\pi_{\phi}^{RL}\) is the learned RL policy, \(\pi^{SFT}(y \mid x)\) is the pre-trained supervised fine-tuning model, \(D_{\text{pretrain}}\) is the pre-training distribution, \(\beta\) is the coefficient for the Kullback–Leibler (KL) reward, and \(\gamma\) is the coefficient for the pre-training loss. These coefficients control the strength of the KL penalty and the pre-training gradient, respectively.

The input and output of LTMs consist of continuous numerical values, presenting a fundamental difference from language models. Additionally, most pure LTMs typically employ loss functions such as Mean Squared Error or Quantile Loss, which are designed to perform multi-step predictions and are incompatible with TD errors. Furthermore, these models do not provide probabilistic outputs, making it impossible to compute probabilities or metrics such as KL divergence. As a result, conventional reinforcement learning frameworks like PPO and RLOO cannot be directly applied to LTMs (except probabilistic time series models).

To address this, we propose TPO, a reinforcement learning framework specifically designed for pure LTMs. TPO adapts the reinforcement learning architecture of LLMs to time series by incorporating a standardized RLHF data format and an advantage function based on REINFORCE. Compared to PPO and RLOO, TPO is more efficient and better suited for time-series models. Unlike PPO, which requires training three models (policy, critic, and reward), and RLOO, which involves training both policy and reward models, TPO trains only a single RL policy model. This design significantly reduces computational complexity and costs. The TPO objective is defined as:

\begin{equation}
\begin{split}
& \text{objective}(\phi) = \\
& \hat{A}\left( y_{1:L}^{RL}, \hat{y}_{L+1:L+H}^{RL} \right) \frac{\pi_{RL}(f_{\phi}^{RL_k}(y_{1:L}^{RL}); \mu_{\phi}^{RL_k}, \sigma_{RL})}{\pi_{RL}(f_{\phi}^{RL_{\text{old}}}(y_{1:L}^{RL}); \mu_{\phi}^{RL_{\text{old}}}, \sigma_{RL})} \\
&\quad + \gamma \left( \alpha \text{MSE} \left( f_{\phi}^{RL}(y_{1:L}^{RL}), \hat{y}_{L+1:L+H}^{\text{chosen}} \right) \right. \\
&\quad \left. + \omega \text{MSE}\left( f_{\phi}^{RL}(y_{1:L}^{RL}), \hat{y}_{L+1:L+H}^{\text{rejected}} \right) \right)
\end{split}
\end{equation}
where \(\hat{A}\left( y_{1:L}^{RL}, \hat{y}_{L+1:L+H}^{RL} \right)\) is the advantage value estimated from the prediction \(\hat{y}_{L+1:L+H}^{RL}\) given the input \(y_{1:L}^{RL}\), \(\pi_{RL}\) represents the probability of the prediction from RL strategy, and \(\frac{\pi_{RL}(f_{\phi}^{RL_k}(y_{1:L}^{RL}); \mu_{\phi}^{RL_k}, \sigma_{RL})}{\pi_{RL}(f_{\phi}^{RL_{\text{old}}}(y_{1:L}^{RL}); \mu_{\phi}^{RL_{\text{old}}}, \sigma_{RL})}\) is the ratio of the new policy to the old policy. The parameter \(k\) is the number of policy update steps in each iteration, \(\gamma\) controls the weight of the time-series loss, and \(\alpha\) and \(\omega\) control the weights of the good and bad prediction losses (in feedback contrast pair), respectively.

Next, we will provide a detailed explanation of the TPO framework, highlighting its key components and discussing the advantages it offers over traditional reinforcement learning methods.

\subsubsection{Feedback Contrast Pair}
For the input, in addition to the original time-series input \(y_{1:L}^{RL} \in \mathbb{R}^L\) from the RLHF dataset, we add a feedback contrast pair which includes a good prediction and a bad prediction \(\left[ \hat{y}_{L+1:L+H}^{\text{chosen}}, \hat{y}_{L+1:L+H}^{\text{rejected}} \right] \in \left[ \mathbb{R}^H, \mathbb{R}^H \right]\). The feedback contrast pair is constructed from a bunch of time-series models built by JD.com's data analysts who have abundant experience on specific forecasting scenarios. That is, these tailored time series models are believed to reflect the cognitive insights, experience, and preferences of these people in forecasting tasks. The objective of fine-tuning on feedback contrast pair is to adjust the LTM model's predictions to be more aligned with the desired outcomes, thereby forcing LTM to learn from the tacit knowledge of human experts.

\subsubsection{Probabilistic Prediction}
Most time series models produce deterministic predictions without probabilities or uncertainties, making it impossible to compute KL divergence or policy probability loss in reinforcement learning. To address this issue, we develop a generic probabilistic prediction component suitable for all time series models (not limited to LTMs). This component assumes that both time series predictions and "good"-"bad" predictions follow a normal distribution \(N(\mu, 1)\). The mean $\mu$ can be directly computed from the model's predicted values, allowing for a rapid generation of prediction probabilities. \(\pi_{RL}\), \(\pi_{\text{chosen}}\), and \(\pi_{\text{rejected}}\) can be derived from this approach.

\subsubsection{Advantage Function}
Since our time series large model directly outputs multi-step forecasts, we avoid using TD error in the RL phase. Instead, we adopt a REINFORCE-inspired approach, using an advantage function to quantify the improvement over a baseline reward. The goal is to fine-tune LTMs to closely align with the original SFT model while encouraging predictions that approach the "good prediction." Larger deviations from the baseline reward yield greater advantages, guiding the model toward outputs that better reflect human expertise.

The advantage function is defined as:
\begin{equation}
\hat{A}\left(y_{1:L}^{RL},{\hat{y}}_{L+1:L+H}^{RL}\right)= R\left(y_{1:L}^{RL},{\hat{y}}_{L+1:L+H}^{RL}\right)-b_{ts}
\end{equation}
where \( R\left(y_{1:L}^{RL},{\hat{y}}_{L+1:L+H}^{RL}\right) \) is the reward function which consists of two parts:
\begin{equation}
\begin{split}
& R\left(y_{1:L}^{RL},{\hat{y}}_{L+1:L+H}^{RL}\right) = r(y_{1:L}^{RL},{\hat{y}}_{L+1:L+H}^{RL}) \\
& -\beta\log{\left(\frac{\pi_{RL}(f_\phi^{RL}\left(y_{1:L}^{RL}\right);\mu_\phi^{RL},\sigma_{RL})}{\pi_{RL}(f_\phi^{SFT}\left(y_{1:L}^{RL}\right);\mu_\phi^{RL},\sigma_{RL})}\right)}
\end{split}
\label{eq_adfunc}
\end{equation}

The first term in Eq. \eqref{eq_adfunc}, \( r\left(y_{1:L}^{RL},{\hat{y}}_{L+1:L+H}^{RL}\right) = \log\left(\pi_{chosen}(f_\phi^{RL}\left(y_{1:L}^{RL}\right);\mu_\phi^{chosen},\sigma_{chosen})\right) \), represents the probability of the current prediction within the distribution of ``good" predictions. Clearly, the higher this probability, the better the prediction will align with the ``good prediction." The second term is the KL divergence, measuring the deviation between the RL model and the SFT model, to prevent excessive divergence during the RL phase. The coefficient \(\beta\) controls the weight of the KL reward term.

In contrast to RLOO's baseline reward, which is derived through online sampling, the baseline reward in our design aims to reduce the variance in model learning. By using the SFT model's prediction (i.e., the ``bad prediction") as the baseline reward, we ensure that the computed difference reflects a gain toward the ``good prediction." This approach avoids the computational costs associated with the sampling method used in RLOO, resulting in more efficient learning:

\begin{equation}
\scalebox{0.85}{$
b_{ts} = \log\left(\pi_{rejected}(f_\phi^{RL}\left(y_{1:L}^{RL}\right);\mu_\phi^{rejected},\sigma_{rejected})\right).
$}
\end{equation}

% \subsubsection{Time Series Loss}
% In the original RL objective, a pre-training loss term is introduced to ensure that the RL model maintains its performance on standard NLP tasks, thereby preventing it from "forgetting" its foundational capabilities. For time series models, Mean Squared Error (MSE) is one of the most widely used error metrics. As such, we incorporate an MSE loss term to enhance the model's forecasting ability during fine-tuning. In constructing this MSE loss, we consider not only the difference between the prediction and the "good" prediction in the feedback contrast pair but also the discrepancy between the prediction and the "bad" prediction. Experimental results demonstrate that this approach effectively reduces overfitting during the fine-tuning process.

\section{Experiments}

\subsection{Construction of Large-scale, High-quality Time Series Datasets}

Scaling up datasets is a proven strategy for improving the performance and generalizability of large time-series models. Time-series data vary in complexity: high-dimensional data often include patterns like trends and seasonality, while low-dimensional data are influenced by random factors such as timing and promotions. By carefully designing datasets, valuable information from diverse sources can be effectively captured.

In this study, we propose a standard for constructing large-scale, high-quality time-series datasets. The dataset is divided into three components: pretraining data, supervised fine-tuning (SFT) data, and reinforcement learning with human feedback (RLHF) data. The pretraining dataset combines JD.com’s proprietary sales data, public datasets, and synthetic data, resulting in a large-scale dataset of 210 billion data points with intricate temporal patterns. Details of the pretraining dataset are provided in Appendix \ref{pretrain_data}.

To adapt the pre-trained model to downstream tasks, we construct fine-tuning datasets tailored to specific scenarios, with further details in Appendix \ref{sft_data}. The RLHF dataset consists of feedback contrast pairs, where each sample includes a good prediction and a bad prediction, as described in Appendix \ref{rlhf_data_jd}.

\subsection{Experiment Setting}

For the JD.com dataset, we evaluate forecasting performance on over 30,000 products, using data from April 2024 onward as the testing period. Forecast Accuracy (FA) is adopted as the evaluation metric, with the current JD.com online algorithm (JD\_ONLINE) serving as a baseline. JD\_ONLINE is an ensemble model combining machine learning methods (e.g., XGBoost) and deep learning models. We compare this with three proposed methods: PCLTM, PCLTM+SFT, and PCLTM+SFT+TPO.

For the public datasets, we evaluate performance using test data from four datasets in the TSLib database: ETTh1, ETTh2, ETTm1, and ETTm2. Metrics used include Mean Squared Error (MSE) and Mean Absolute Error (MAE), with detailed calculation formulas provided in Appendix \ref{formula_eval}. Comparisons are made against the state-of-the-art finetuned LTM (GPT4TS) and five advanced full-shot time-series models: PatchTST, Autoformer, ITransformer, DLinear, and Informer. Model configurations and training details are detailed in Appendix \ref{model_conf}.

\subsection{Results}

\subsubsection{Results on JD.com's Dataset}
We use a model size of 300M for efficiency to compare our method and the baseline (JD\_ONLINE). A detailed comparison of different model sizes is provided later in Section \ref{subsec_ablation}. Table \ref{tab_model_accuracy_improvement} shows the comparison, and the results indicate that our method consistently outperforms JD\_ONLINE. Specifically, the incremental improvements from PCLTM, SFT, and TPO demonstrate that each component contributes to the overall performance, validating the effectiveness of our framework. The greatest improvement was achieved with PCLTM+SFT+TPO, resulting in a 14.77\% increase in prediction accuracy.

\begin{table}[h]
\caption{Model and Relative Accuracy Improvement}
\label{tab_model_accuracy_improvement}
\begin{center}
\begin{small}
\begin{sc}
\begin{tabular}{cc}
\toprule
Model & \makecell{Relative \\ Accuracy \\ Improvement} \\ 
\midrule
JD\_ONLINE & - \\
PCLTM(300M)+ZERO SHOT & 9.47\% \\
PCLTM(300M)+SFT & 10.98\% \\
PCLTM(300M)+SFT+TPO & 14.77\% \\
\bottomrule
\end{tabular}
\end{sc}
\end{small}
\end{center}
\vskip -0.1in
\end{table}

Table \ref{tab_scenario_improvement} presents the accuracy improvements of our method across different scenarios. Notable performance gains are observed in the best-selling, seasonal, long-tail, and new product scenarios. For the PCLTM+SFT+TPO model, the relative optimization ranges from 3.99\% to 29.51\%, with the most substantial improvement seen in the long-tail product scenario.

\begin{table}[h]
\caption{Performance Improvement Across Different Scenarios}
\label{tab_scenario_improvement}
\begin{center}
\begin{scriptsize}
\begin{sc}
\resizebox{0.5\textwidth}{!}{
\begin{tabular}{llc}
\toprule
Scenario & Model & Relative Accuracy Improvement \\ 
\midrule
\multirow{4}{*}{Bestsellers} & JD\_ONLINE & - \\
                              & PCLTM(300m)+ZERO SHOT & 16.86\% \\
                              & PCLTM(300m)+SFT & 17.99\% \\
                              & PCLTM(300m)+SFT+TPO & 20.08\% \\
\midrule
\multirow{4}{*}{Seasonal Items} & JD\_ONLINE & - \\
                                 & PCLTM(300m)+ZERO SHOT & 6.25\% \\
                                 & PCLTM(300m)+SFT & 9.62\% \\
                                 & PCLTM(300m)+SFT+TPO & 14.42\% \\
\midrule
\multirow{4}{*}{Long-tail Products} & JD\_ONLINE & - \\
                                     & PCLTM(300m)+ZERO SHOT & 7.45\% \\
                                     & PCLTM(300m)+SFT & 9.46\% \\
                                     & PCLTM(300m)+SFT+TPO & 29.51\% \\
\midrule
\multirow{4}{*}{New Products} & JD\_ONLINE & - \\
                              & PCLTM(300m)+ZERO SHOT & -0.84\% \\
                              & PCLTM(300m)+SFT & 0.84\% \\
                              & PCLTM(300m)+SFT+TPO & 3.99\% \\
\bottomrule
\end{tabular}
}
\end{sc}
\end{scriptsize}
\end{center}
\vskip -0.1in
\end{table}

To visually present the prediction performance of our method, we display the prediction results of both our method and JD\_Online across different products in Appendix \ref{CASE_detail_online}. To demonstrate the effects of SFT and TPO, we also show the predictions of PCLTM, PCLTM+SFT, and PCLTM+SFT+TPO in Appendix \ref{CASE_detail_tpo}.

\textbf{Deployment on JD.com}. In December 2024, we successfully deployed our proposed LTM on JD.com. The model is currently being used for automated replenishment across 20,000 SKUs. Compared to the previous online forecasting system, the model has delivered a substantial 33.21\% improvement in prediction accuracy.

\subsubsection{Results on Public Datasets}
On several public datasets, our method, in a zero-shot setting, performs comparably to or slightly better than the current state-of-the-art fine-tuned LTM (e.g., GPT4TS) and leading deep learning time series models. The performance metrics, including MSE and MAE, are presented in Table \ref{tab_mse_mae} and more detail in Appendix\ref{mae_mse_detail}, with the best results marked in bold and the second-best results underlined. The results indicate a progressive improvement as we incorporate SFT and SFT+TPO, with our model consistently ranking among the top 2 in most cases. This demonstrates the superiority of our proposed TimeHF framework.
\begin{table*}[h!]
\caption{Performance Comparison of Our Method and Baselines Using MSE And MAE}
\label{tab_mse_mae}
\begin{center}
\begin{small}
\begin{sc}
\resizebox{2\columnwidth}{!}{
\begin{tabular}{cc|ccc|c|ccccc}
\toprule
Dataset & Metrics & \multicolumn{3}{c}{PCLTM(300M)} & GPT4TS & PatchTST & Autoformer & iTransformer & DLinear & Informer \\
 &  & ZERO SHOT & SFT & SFT+TPO & SFT &  \multicolumn{5}{c}{FULL SHOT}  \\
\midrule
etth1 & mse & 0.3999 & 0.3645 & \textbf{0.3503} & 0.3697 & \underline{0.3625} & 0.4452 & 0.3746 & 0.3826 & 0.7144\\
 & mae & 0.3994 & 0.3924 & \textbf{0.3819} & 0.4000 & \underline{0.3851} & 0.4536 & 0.3818 & 0.3921 & 0.7934 \\
\midrule
etth2 & mse & 0.2262 & \underline{0.2199} & \textbf{0.2175} & 0.2419 & 0.2347 & 0.2839 & 0.2623 & 0.2967 & 1.1863 \\
 & mae & 0.3028 & \underline{0.2986} & \textbf{0.2956} & 0.3170 & 0.3011 & 0.3502 & 0.3321 & 0.3714 & 0.9168 \\
\midrule
ettm1 & mse & 0.2834 & 0.2692 & \textbf{0.2606} & \underline{0.2621} & 0.2915 & 0.4567 & 0.3226 & 0.3388 & 0.7000 \\
 & mae & 0.3355 & 0.3257 & \underline{0.3216} & \textbf{0.3212} & 0.3550 & 0.4495 & 0.3565 & 0.3614 & 0.9571 \\
\midrule
ettm2 & mse & 0.1417 & \underline{0.1295} & \textbf{0.1285} & 0.1468 & 0.1573 & 0.1968 & 0.1665 & 0.1778 & 0.5592 \\
 & mae & 0.2319 & \underline{0.2214} & \textbf{0.2182} & 0.2396 & 0.2464 & 0.2976 & 0.2615 & 0.2910 & 0.4749 \\
\bottomrule
\end{tabular}
}
\end{sc}
\end{small}
\end{center}
\vskip -0.1in
\end{table*}

\subsection{Ablation Analysis}
\label{subsec_ablation}

We conduct a detailed ablation study on our framework and the relative accuracy improvement for the model without each component is shown in Table \ref{tab_accuracy_improvement}. Overall, each component we propose contributes positively to the overall performance. Among them, TPO provides the most significant improvement; removing TPO greatly affects the model's predictive accuracy. The second most impactful component is the temporal positional encoding, followed by PatchConv and SFT.

\begin{table}[h!]
\caption{Ablation Analysis: Relative Accuracy Improvement of Our Framework with the Exclusion of Each Component}
\label{tab_accuracy_improvement}
\begin{center}
\begin{small}
\begin{sc}
\begin{tabular}{lc}
\toprule
Model Configuration & \makecell{Relative \\ Accuracy \\ Improvement} \\ 
\midrule
PCLTM(300M)+SFT+TPO & \textbf{14.77\%} \\
w/o PatchConv & 12.87\% \\
w/o ROPE & 11.87\% \\
w/o SFT & 13.47\% \\
w/o TPO & 10.98\% \\
\bottomrule
\end{tabular}
\end{sc}
\end{small}
\end{center}
\vskip -0.1in
\end{table}

We also compare different RLHF methods to investigate the effect of TPO. Table \ref{tab_rl_comp} shows the relative accuracy improvement of PCLTM(300M)+SFT compared with different RLHF methods or TPO under various parameter settings, along with training efficiency. We find that TPO demonstrates an obvious advantage (12.34\%-14.77\%) over other mainstream RL frameworks in time series tasks. Furthermore, removing either the MSE\_loss or the MSE\_rejected\_loss from the TPO objective function results in a noticeable decline in relative accuracy. This highlights the necessity of the proposed feedback contrast pair. These pairs allow LTMs to learn from the rich forecasting expertise of JD.com’s business experts, leading to further improvements in accuracy.

In terms of training efficiency, our TPO method shows a significant improvement compared to other common RL methods. While DPO demonstrates the highest efficiency, TPO closely follows with only a marginal difference of 0.02 it/s. In terms of predictive performance, however, TPO significantly surpasses DPO by over 3.5 percentage points.

\begin{table}[h!]
\caption{Relative Accuracy Improvement and Training Efficiency of RLHF methods}
\label{tab_rl_comp}
\begin{center}
\resizebox{\columnwidth}{!}{ % 缩放表格以适应单列宽度
\begin{tabular}{lcc}
\toprule
Model/Configuration & \makecell{Relative \\ Accuracy \\ Improvement} & \makecell{Training \\ Efficiency \\ (itr/s, Single GPU)} \\ 
\midrule
PCLTM(300m)+SFT+PPO & 10.98\% & 0.4 \\
PCLTM(300m)+SFT+DPO & 11.15\% & \textbf{1.86} \\
PCLTM(300m)+SFT+RLOO & 11.36\% & 1.05 \\ 
PCLTM(300m)+SFT+TPO ($\gamma=0$) & 12.34\% & 1.75 \\
PCLTM(300m)+SFT+TPO ($\omega=0$) & \underline{13.23\%} & \underline{1.84} \\
PCLTM(300m)+SFT+TPO & \textbf{14.77\%} & \underline{1.84} \\
\bottomrule
\end{tabular}
}
\end{center}
\vskip -0.1in
\end{table}

% \begin{table*}[h!]
% \caption{Relative Accuracy Improvement and Training Efficiency of RLHF methods}
% \label{tab_rl_comp}
% \begin{center}
% \resizebox{0.6\textwidth}{!}{ % 按比例缩放表格
% \begin{tabular}{lcc}
% \toprule
% Model/Configuration & \makecell{Relative \\ Accuracy \\ Improvement} & \makecell{Training \\ Efficiency \\ (itr/s, Single GPU)} \\ 
% \midrule
% PCLTM(300m)+SFT+PPO & 10.98\% & 0.4 \\
% PCLTM(300m)+SFT+DPO & 11.15\% & \textbf{1.86} \\
% PCLTM(300m)+SFT+RLOO & 11.36\% & 1.05 \\ 
% PCLTM(300m)+SFT+TPO ($\gamma=0$) & 12.34\% & 1.75 \\
% PCLTM(300m)+SFT+TPO ($\omega=0$) & \underline{13.23\%} & \underline{1.84} \\
% PCLTM(300m)+SFT+TPO & \textbf{14.77\%} & \underline{1.84} \\
% \bottomrule
% \end{tabular}
% }
% \end{center}
% \vskip -0.1in
% \end{table*}

\subsection{Hyperparameter Analysis}

We calculate the metrics for PCLTMs with different parameter counts and present them in Table \ref{tab_model_size}. As shown, the accuracy improves with the increase in the number of parameters, further validating the scaling law. This suggests that for LTM, ultra-large amounts of parameters are a key factor in improving prediction performance. As mentioned earlier, large-scale models require vast amounts of high-quality time series data for effective training and fine-tuning, which also highlights the necessity of our carefully designed pretraining, SFT, and RLHF datasets.

\begin{table}[h!]
\caption{Model Size and Relative Accuracy Improvement}
\label{tab_model_size}
\begin{center}
\begin{small}
\begin{sc}
\begin{tabular}{lc}
\toprule
Model Size & Relative Accuracy Improvement \\ 
\midrule
PCLTM(300M) & 9.47\% \\
PCLTM(1.6B) & 10.22\% \\
PCLTM(6B)   & 10.68\% \\ 
\bottomrule
\end{tabular}
\end{sc}
\end{small}
\end{center}
\end{table}

\begin{figure}[ht!]
\begin{center}
\centerline{\includegraphics[width=\columnwidth]{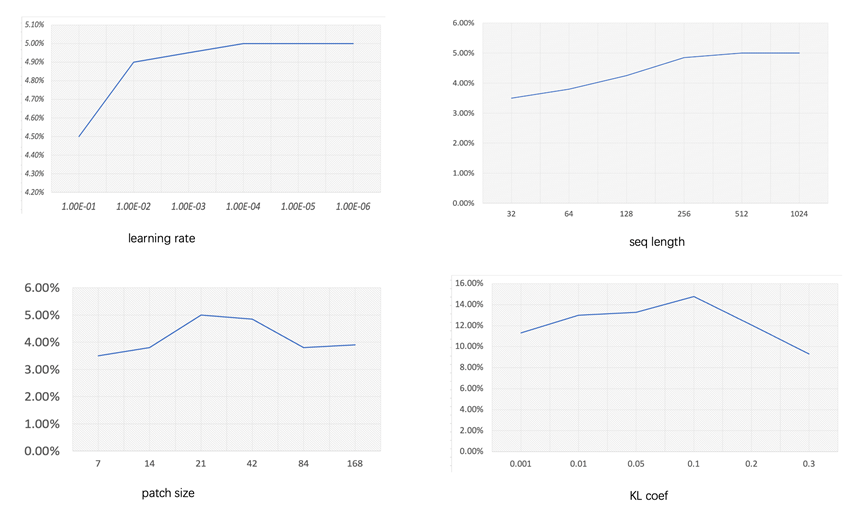}}
\caption{Impact of Hyperparameters on Accuracy Improvement}
\label{fig_hyper}
\end{center}
\end{figure}

We analyze the model performance under different learning rates, sequence lengths, patch sizes, and KL coefficients, as shown in Figure \ref{fig_hyper}. Proper hyperparameters can effectively improve prediction accuracy.

We tested the effects of different settings for the hyperparameters \(\alpha\) and \(\omega\) in the MSE loss of the TPO objective function. Table \ref{tab_alpha_omega_accuracy} shows that the prediction accuracy improves when the parameters are set to (0.9, 0.1) and (0.8, 0.2).

\begin{table}[h!]
\caption{Relative Accuracy Improvement for Different $(\alpha, \omega)$ Values in TPO Objective}
\label{tab_alpha_omega_accuracy}
\begin{center}
\begin{small}
\begin{sc}
\begin{tabular}{cc}
\toprule
$(\alpha, \omega)$ & Relative Accuracy Improvement \\ 
\midrule
(1.0, 0.0) & 13.23\% \\
(0.9, 0.1) & 14.29\% \\
(0.8, 0.2) & 14.77\% \\
(0.7, 0.3) & 13.76\% \\
(0.6, 0.4) & 11.40\% \\
(0.5, 0.5) & 10.84\% \\
(0.4, 0.6) & 9.89\% \\
\bottomrule
\end{tabular}
\end{sc}
\end{small}
\end{center}
\end{table}

\section{Conclusion}
In this paper, we present a novel training framework for large-scale time series models (PCLTM+SFT+TPO). PCLTM is a base LTM, and it is the first time series model at the billion-parameter scale. Its zero-shot performance outperforms the current state-of-the-art fine-tuned large models, such as GPT4TS, as well as fully-supervised forecasting models across various time series datasets. We also propose TPO, an RLHF framework that is tailored for pure LTMs. Our experiments show that TPO consistently outperforms existing RLHF frameworks (such as PPO and RLOO) in predictive performance. Our methodology has practical impacts. The proposed method has already been deployed on JD.com's supply chain system for an automated replenishment process.

Our work provides a new perspective for building smarter LTMs, specifically by incorporating human expert feedback into the training process using RLHF methods. We hope that future research will further explore this direction and continue to advance the integration of human knowledge into time series modeling.

\section*{Impact Statement}
This work introduces a novel approach to time series modeling by combining pretraining, supervised fine-tuning, and reinforcement learning with human feedback. Our proposed framework marks a significant advancement in predictive accuracy for diverse time series tasks, consistently outperforming state-of-the-art methods in both zero-shot and fine-tuned settings. Notably, it demonstrates remarkable improvements in sales prediction across a variety of products, showcasing its versatility and robustness.

The practical implications of this research are profound. In the supply chain domain, even marginal improvements in forecasting accuracy can yield substantial financial gains. At JD.com, for example, a 10\% relative improvement in prediction accuracy translates into annual cost savings of tens of millions of yuan and sales growth worth billions or even hundreds of billions of yuan. The model introduced in this paper has already been fully deployed at JD.com, achieving over a 30\% enhancement in forecasting accuracy and delivering extraordinary economic benefits, reinforcing its real-world impact.

Beyond these tangible results, this research lays the foundation for integrating expert feedback into time series forecasting, paving the way for scalable and adaptable solutions across various domains. We believe this framework offers immense potential for future applications, and we anticipate that subsequent studies will build upon these findings to unlock even greater opportunities in both academia and industry.

\section*{Acknowledgements}
The authors would like to extend their sincere gratitude for the support received from Computing Department
 of JD.com, with particular thanks to Zhen Chen, XiaoKun Zhu, Zhaolong Xing, Yang Pei, Qian Yu.

\bibliography{main}
\bibliographystyle{plainnat}

%%%%%%%%%%%%%%%%%%%%%%%%%%%%%%%%%%%%%%%%%%%%%%%%%%%%%%%%%%%%%%%%%%%%%%%%%%%%%%%
%%%%%%%%%%%%%%%%%%%%%%%%%%%%%%%%%%%%%%%%%%%%%%%%%%%%%%%%%%%%%%%%%%%%%%%%%%%%%%%
% APPENDIX
%%%%%%%%%%%%%%%%%%%%%%%%%%%%%%%%%%%%%%%%%%%%%%%%%%%%%%%%%%%%%%%%%%%%%%%%%%%%%%%
%%%%%%%%%%%%%%%%%%%%%%%%%%%%%%%%%%%%%%%%%%%%%%%%%%%%%%%%%%%%%%%%%%%%%%%%%%%%%%%
\newpage
\appendix
\onecolumn
\section{Dataset Detail  }
\subsection{Pretrain Dataset Detail}\label{pretrain_data}
The entire dataset consists of JD.com’s proprietary sales data, publicly available datasets, and synthetic data. The JD dataset includes extensive sales information, such as lifecycle, promotions, marketing activities, unexpected events, and inventory status, adding complexity to the data. Public datasets cover industries like electricity, solar energy, weather, and transportation, featuring strong regularity with common time series patterns like seasonality and trends. Synthetic datasets are created using high-dimensional aggregation and interpretable methods, ensuring high accuracy and enabling the model to effectively learn the knowledge.the data have totaling approximately 488 billion observational data points. The data set undergoes a series of processing steps, including labeling, quality filtering, deduplication, diversity ranking, and data balancing, resulting in a final pretraining dataset with approximately 210 billion observational points
\textbf{JD.com Dataset.} The majority of the data is sourced from JD.com’s sales data, covering various categories such as food and clothing over the past three years. This data contains approximately 382B observational points.

\textbf{Public Datasets.} We also incorporate data from the Monash Time Series Database and the TSLib Database, expanding the training samples through random segments. This contributes around 8B observational points.

\textbf{Synthetic Datasets.} We use two data augmentation methods to construct synthetic data: (1) Interpretable component prediction based on JD.com and public dataset time series. We perform component forecasting on historical time series, with components including baseline, seasonality, promotions, and holidays. The presence of components helps the model learn different time series characteristics more distinctly. (2) high-dimensional aggregated, time series are aggregated across different horizons (e.g. week, month) and different dimensions (e.g. category, brand, region), enriching the dataset's diversity and heterogeneity. This dataset contains approximately 98 billion observational points. 

\textbf{Pretraining Data Processing.} The base data undergoes the following five steps, with steps 1-4 specifically applied to the non-public data: (1) Labeling: Each sample in the non-public datasets is labeled with some metrics, such as series length, average daily sales, and zero-sales proportion. These labels are used to describe the characteristics of each time series. (2) Quality Filtering: Time series are evaluated for quality based on the labeled features, and those with excessively short lengths or polluted sales are removed, improving the overall data quality. (3) Deduplication: The data is randomly grouped and clustered by time series. A subset of samples within each cluster is retained to ensure diversity without redundancy. (4) Diversity Ranking: The data is reordered based on time series labels, ensuring each batch contains diverse time series with varying characteristics. (5) Data Balancing:
We set different proportions for each data source: 20\% synthetic data, 4\% public datasets, and 76\% JD.com data. We also balance aggregated time series (30\%) with regular time series (70\%). Resampling is performed to ensure the final dataset adheres to these balance configurations. After evaluating the results from multiple data ratios on the traditional time series models, the optimal ratio configuration is selected for final integration into the pretraining dataset.

\subsection{SFT Dataset Detail}\label{sft_data}
To better adapt the pre-trained model for downstream tasks, we construct fine-tuning datasets tailored to specific scenarios.

For JD.com's scenario, we divide the data based on product types into four categories: best-sellers, seasonal items, long-tail products, and newly launched items. Fine-tuning datasets were constructed for each category.
\begin{itemize}
    \item Best-sellers: Products with relatively high sales but lower quantity in terms of total product volume.
    \item Seasonal items: Products with significant seasonality, e.g., those popular in summer but less so in winter, or vice versa.
    \item Long-tail products: Items with low average daily sales and a high proportion of zero sales.
    \item Newly launched items: Products with less than six months of historical sales data.
\end{itemize}
We select data ratios best suited for each category. For instance, in the best-seller category where the time series is more regular and stable, we reduce the proportion of synthetic data and increase the proportion of real data. The final fine-tuning dataset for JD.com's scenario has an average of approximately 4B data points.
For the public data scenario, we finetune the model using training data from the four datasets in the TSLib database: ETTh1, ETTh2, ETTm1, and ETTm2.

\subsection{RLHF Dataset Detail}\label{rlhf_data_jd}

In the RLHF fine-tuning dataset, each sample is a feedback contrast pair consisting of a good prediction and a bad prediction. For example, given historical data for the past 6 days, the goal is to forecast the next 3 days. The sample includes the 6 historical observations, with both the good and bad predictions being the 3 forecasted points for the future.

In the JD.com data scenario, business experts from various sectors, such as large supermarkets, books, and 3C electronics, provide forecasts based on their experience and expertise. These forecasts are paired with predictions from the large model and evaluated by the experts. The predictions with smaller forecasting errors are labeled as good predictions, while those with larger errors are labeled as bad predictions. A total of 4532 pairs were collected for this purpose as shown in Table \ref{tab_sector_sample}.

For public datasets, we apply multiple prediction methods on the validation set of public datasets, including the outputs of the large model and comparative time series models (PatchTST, iTransformer, Autoformer). Predictions are manually labeled by algorithm engineers to form feedback contrast pairs. The number of samples for each public dataset is shown in Table \ref{tab_pub_sample}.

\begin{table}[h]
\caption{Number of Feedback Contrast Pairs for JD.com's sectors}
\label{tab_sector_sample}
\vskip 0.15in
\begin{center}
\begin{small}
\begin{sc}
\begin{tabular}{cc}
\toprule
Sector        & \#Feedback Contrast Pairs \\ 
\midrule
Home Appliances & 634      \\
Digital Products   & 310      \\
Books           & 786      \\
Fashion         & 47       \\
Pharmaceuticals & 531      \\
Automobiles     & 513      \\
Food \& Lifestyle  & 1,711     \\
Total           & 4,532     \\
\bottomrule
\end{tabular}
\end{sc}
\end{small}
\end{center}
\vskip -0.1in
\end{table}

\begin{table}[h]
\caption{Number of Feedback Contrast Pairs for Public Datasets}
\label{tab_pub_sample}
\vskip 0.15in
\begin{center}
\begin{small}
\begin{sc}
\begin{tabular}{cc}
\toprule
Public Dataset & \#Feedback Contrast Pairs \\ 
\midrule
ETTh1           & 2,683     \\
ETTh2           & 1,110     \\
ETTm1           & 4,218     \\
ETTm2           & 2,574     \\
\bottomrule
\end{tabular}
\end{sc}
\end{small}
\end{center}
\vskip -0.1in
\end{table}

\section{Evaluation Detail  }
\subsection{Evaluation formula}\label{formula_eval}
\begin{equation}
\text{FA} = 1 - \frac{\left|\sum_{h=lt}^{lt+bp} y_{l+h} - \sum_{h=lt}^{lt+bp} \hat{y}_{l+h}\right|}{\sum_{h=lt}^{lt+bp} \left|y_{l+h}\right|}
\end{equation}
\begin{equation}
\text{MSE} = \frac{1}{H} \sum_{h=1}^{H} \left(y_{l+h} - \hat{y}_{l+h}\right)^2
\end{equation}
\begin{equation}
\text{MAE} = \frac{1}{H} \sum_{h=1}^{H} \left|y_{l+h} - \hat{y}_{l+h}\right|
\end{equation}
where \( H \) represents the forecasting horizon, i.e., the number of steps ahead for prediction. \( y_h \) and \( \hat{y}_h \) denote the actual and predicted values at step \( h \), respectively. \( lt \) and \( bp \) refer to the delivery lead time and the procurement cycle, respectively.

\subsection{Model configuration and training details}\label{model_conf}
Model configuration and training details are shown in Table \ref{tab_model_config} and \ref{tab_training_details}.

\begin{table}[h]
\caption{Model Configuration}
\label{tab_model_config}
\vskip 0.15in
\begin{center}
\begin{small}
\begin{sc}
\begin{tabular}{lccc}
\toprule
Model Size & 300M & 1.6B & 6B \\
\midrule
seq\_len & 512 & 512 & 512 \\
patch\_len & 21 & 21 & 21 \\
layers & 24 & 32 & 20 \\
dmodel & 1024 & 2048 & 5120 \\
head & 16 & 16 & 16 \\
\bottomrule
\end{tabular}
\end{sc}
\end{small}
\end{center}
\vskip -0.1in
\end{table}

\begin{table}[h]
\caption{Training Details}
\label{tab_training_details}
\vskip 0.15in
\begin{center}
\begin{small}
\begin{sc}
\begin{tabular}{lccc}
\toprule
Model Size & 300M & 1.6B & 6B \\
\midrule
\#GPUs & 4 & 8 & 12 \\ 
batch size & 4*32 & 8*4 & 12*4 \\
learning rate & $1\times 10^{-4}$ & $1\times 10^{-4}$ & $3\times 10^{-4}$ \\
deep speed & \makecell{None\\ or stage1} & stage1 & stage2 \\
\bottomrule
\end{tabular}
\end{sc}
\end{small}
\end{center}
\vskip -0.1in
\end{table}

\section{Performance Comparison of Our Method and Baselines  }\label{mae_mse_detail}

\begin{table*}[htbp!]
\caption{Performance Comparison of Our Method and Baselines Using MSE}
\label{tab_mse}
\vskip 0.15in
\begin{center}
\begin{small}
\begin{sc}
\resizebox{\columnwidth}{!}{
\begin{tabular}{cc|ccc|c|ccccc}
\toprule
Dataset & Horizon & \multicolumn{3}{c}{PCLTM(300M)} & GPT4TS & PatchTST & Autoformer & iTransformer & DLinear & Informer \\
 &  & ZERO SHOT & SFT & SFT+TPO & SFT &  \multicolumn{5}{c}{FULL SHOT}  \\
\midrule
etth1 & 21 & 0.4281 & 0.3548 & 0.3291 & 0.3622 & \underline{0.3110} & 0.4414 & \textbf{0.3082} & 0.3792 & 0.6198 \\
 & 96 & \underline{0.3716} & 0.3741 & \textbf{0.3714} & 0.3771 & 0.4140 & 0.4490 & 0.4410 & 0.3860 & 0.8090 \\
 & avg & 0.3999 & 0.3645 & \textbf{0.3503} & 0.3697 & \underline{0.3625} & 0.4452 & 0.3746 & 0.3826 & 0.7144 \\
\midrule
etth2 & 21 & 0.1754 & \underline{0.1647} & \textbf{0.1609} & 0.1943 & 0.1674 & 0.2218 & 0.2276 & 0.2604 & 1.1418 \\
 & 96 & 0.2769 & \underline{0.2751} & \textbf{0.2741} & 0.2895 & 0.3020 & 0.3460 & 0.2970 & 0.3330 & 1.2308 \\
 & avg & 0.2262 & \underline{0.2199} & \textbf{0.2175} & 0.2419 & 0.2347 & 0.2839 & 0.2623 & 0.2967 & 1.1863 \\
\midrule
ettm1 & 21 & 0.2248 & \underline{0.2150} & \textbf{0.2002} & 0.2270 & 0.2539 & 0.4083 & 0.3111 & 0.3326 & 0.6909 \\
 & 96 & 0.3419 & 0.3233 & \underline{0.3210} & \textbf{0.2972} & 0.3290 & 0.5050 & 0.3340 & 0.3450 & 0.7090 \\
 & avg & 0.2834 & 0.2692 & \textbf{0.2606} & \underline{0.2621} & 0.2915 & 0.4567 & 0.3226 & 0.3388 & 0.7000 \\
\midrule
ettm2 & 21 & 0.1046 & \underline{0.0911} & \textbf{0.0904} & 0.1206 & 0.1396 & 0.1385 & 0.1529 & 0.1626 & 0.5092 \\
 & 96 & 0.1787 & \underline{0.1678} & \textbf{0.1665} & 0.1730 & 0.1750 & 0.2550 & 0.1800 & 0.1930 & 0.6092 \\
 & avg & 0.1417 & \underline{0.1295} & \textbf{0.1285} & 0.1468 & 0.1573 & 0.1968 & 0.1665 & 0.1778 & 0.5592 \\
\bottomrule
\end{tabular}
}
\end{sc}
\end{small}
\end{center}
\vskip -0.1in
\end{table*}

\begin{table*}[h!]
\caption{Performance Comparison of Our Method and Baselines Using MAE}
\label{tab_mae}
\vskip 0.15in
\begin{center}
\begin{small}
\begin{sc}
\resizebox{\columnwidth}{!}{
\begin{tabular}{cc|ccc|c|ccccc}
\toprule
Dataset & Horizon & \multicolumn{3}{c}{PCLTM(300M)} & GPT4TS & PatchTST & Autoformer & iTransformer & DLinear & Informer \\
 &  & ZERO SHOT & SFT & SFT+TPO & SFT &  \multicolumn{5}{c}{FULL SHOT}  \\
\midrule
etth1 & 21 & 0.4028 & 0.3878 & 0.3680 & 0.3973 & \textbf{0.3511} & 0.4482 & \underline{0.3586} & 0.3842 & 0.7488 \\
 & 96 & \underline{0.3960} & 0.3970 & \textbf{0.3958} & 0.4026 & 0.4190 & 0.4590 & 0.4050 & 0.4000 & 0.8380 \\
 & avg & 0.3994 & 0.3924 & \textbf{0.3819} & 0.4000 & \underline{0.3851} & 0.4536 & 0.3818 & 0.3921 & 0.7934 \\
\midrule
etth2 & 21 & 0.2687 & 0.2612 & \underline{0.2563} & 0.2860 & \textbf{0.2541} & 0.3124 & 0.3152 & 0.3557 & 0.8636 \\
 & 96 & 0.3368 & \underline{0.3359} & \textbf{0.3348} & 0.3479 & 0.3480 & 0.3880 & 0.3490 & 0.3870 & 0.9700 \\
 & avg & 0.3028 & \underline{0.2986} & \textbf{0.2956} & 0.3170 & 0.3011 & 0.3502 & 0.3321 & 0.3714 & 0.9168 \\
\midrule
ettm1 & 21 & 0.2939 & \underline{0.2889} & \textbf{0.2829} & 0.2925 & 0.3429 & 0.4239 & 0.3450 & 0.3508 & 0.9341 \\
 & 96 & 0.3770 & 0.3625 & \underline{0.3602} & \textbf{0.3499} & 0.3670 & 0.4750 & 0.3680 & 0.3720 & 0.9800 \\
 & avg & 0.3355 & 0.3257 & \underline{0.3216} & \textbf{0.3212} & 0.3550 & 0.4495 & 0.3565 & 0.3614 & 0.9571 \\
\midrule
ettm2 & 21 & 0.1980 & \underline{0.1876} & \textbf{0.1860} & 0.2171 & 0.2337 & 0.2561 & 0.2590 & 0.2900 & 0.4197 \\
 & 96 & 0.2658 & \underline{0.2551} & \textbf{0.2504} & 0.2620 & 0.2590 & 0.3390 & 0.2640 & 0.2920 & 0.5300 \\
 & avg & 0.2319 & \underline{0.2214} & \textbf{0.2182} & 0.2396 & 0.2464 & 0.2976 & 0.2615 & 0.2910 & 0.4749 \\
\bottomrule
\end{tabular}
}
\end{sc}
\end{small}
\end{center}
\vskip -0.1in
\end{table*}

\section{Case study}

\subsection{ the prediction results of both our
method and JD Online across different products }\label{CASE_detail_online}
\begin{figure}[ht]
\vskip 0.2in
\begin{center}
\centerline{\includegraphics[width=1\columnwidth]{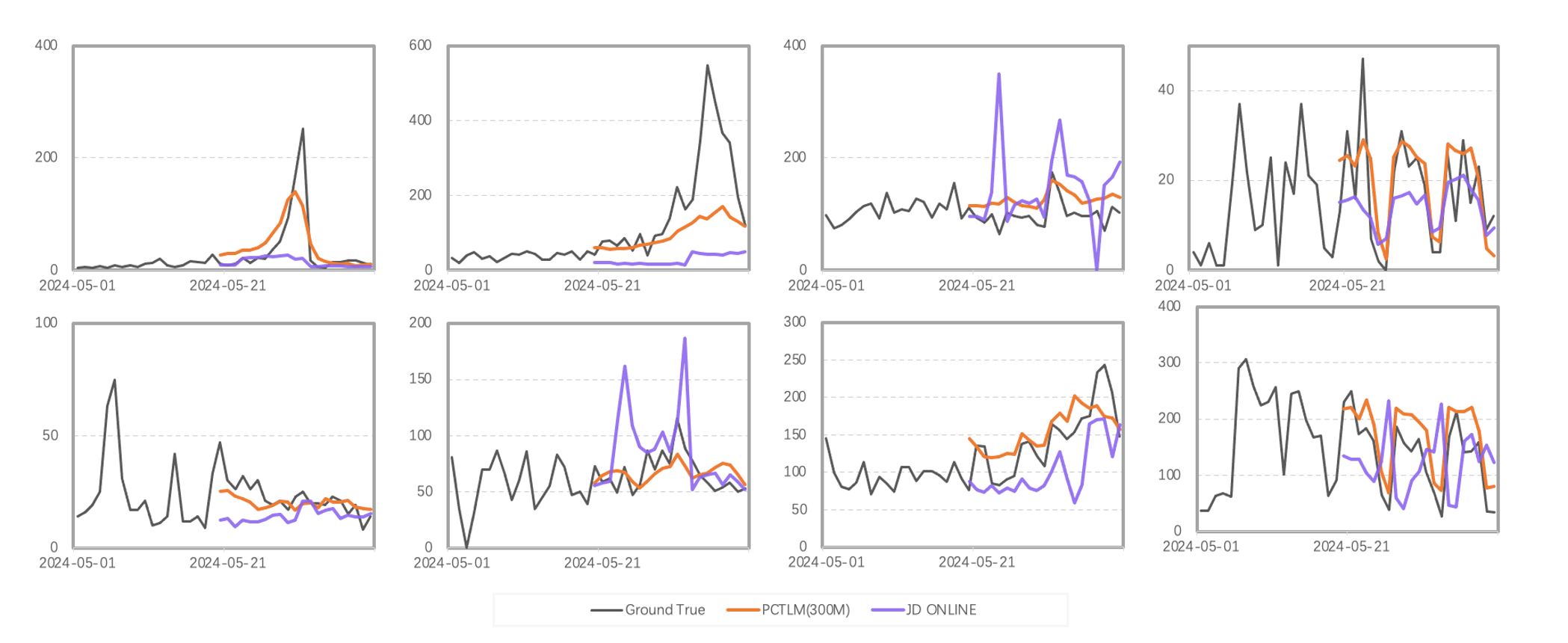}}
\caption{Prediction Comparison Between Our Method and JD\_Online Across Different Products}
\label{fig_jd_comp}
\end{center}
\vskip -0.2in
\end{figure}

\subsection{ Prediction Results of PCLTM,
PCLTM+SFT, and PCLTM+SFT+TPO }\label{CASE_detail_tpo}

\begin{figure}[htbp!]
\vskip 0.2in
\begin{center}
\centerline{\includegraphics[width=\columnwidth]{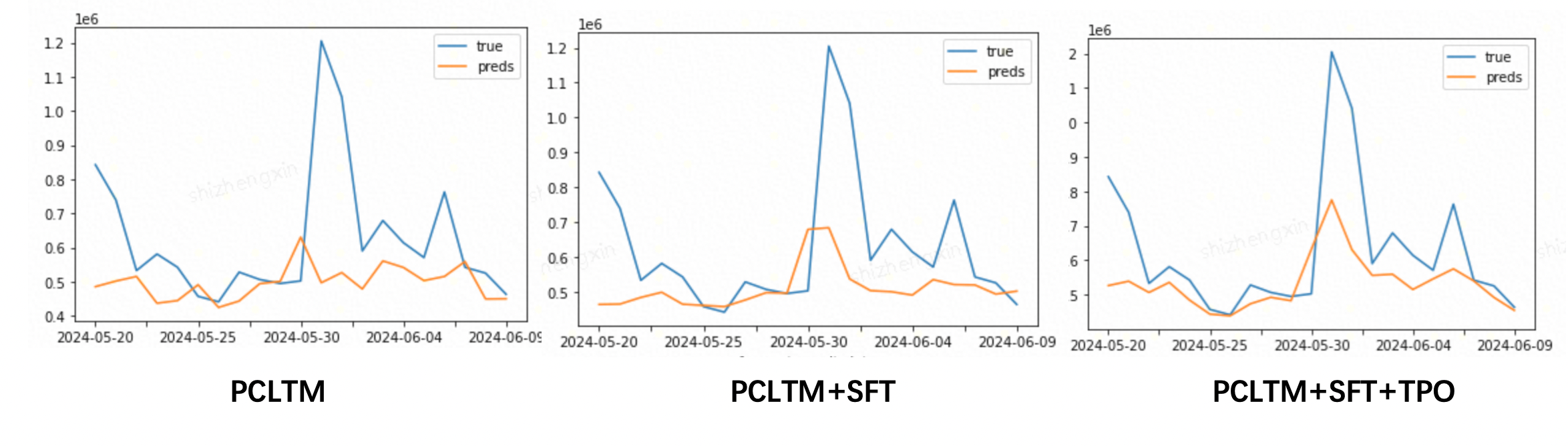}}
\caption{Impact of SFT and TPO: Prediction Results of PCLTM, PCLTM+SFT, and PCLTM+SFT+TPO}
\label{fig_tpo_comp}
\end{center}
\vskip -0.2in
\end{figure}

%%%%%%%%%%%%%%%%%%%%%%%%%%%%%%%%%%%%%%%%%%%%%%%%%%%%%%%%%%%%%%%%%%%%%%%%%%%%%%%
%%%%%%%%%%%%%%%%%%%%%%%%%%%%%%%%%%%%%%%%%%%%%%%%%%%%%%%%%%%%%%%%%%%%%%%%%%%%%%%

\end{document}